\documentclass[a4paper]{article}
\usepackage{RR}
\usepackage[utf8]{inputenc}

\usepackage{hyperref}
\usepackage{amsmath}
\usepackage{amssymb}
\usepackage{enumerate}

\usepackage{amsmath}
\usepackage{amssymb}
\usepackage{enumerate}

\usepackage{graphicx}

\newtheorem{de}{Definition}
\newtheorem{theo}{Theorem}
\newtheorem{prop}[theo]{Proposition}    
\newtheorem{corrol}[theo]{Corollary}
\def\d{\mathbf{d}}

\usepackage{bibunits}
\makeatletter
\def\neutralisetitre{\def\section{\@ifstar\@gobble\@gobble}}
\makeatother
\RRdate{February 2012} 

\RRauthor{
Hachem Kadri\thanks{Sequel Team, INRIA Lille. E-mail: hachem.kadri@inria.fr}%
\and
Alain Rakotomamonjy\thanks{LITIS, Universit\'e de Rouen. E-mail: alain.rakotomamonjy@insa-rouen.fr}%
  \and
Francis Bach\thanks{Sierra Team/INRIA, Ecole Normale Sup\'erieure. E-mail: francis.bach@inria.fr}%
\and
Philippe Preux\thanks{Sequel/INRIA-Lille, LIFL/CNRS. E-mail: philippe.preux@inria.fr}%
} 

\authorhead{Kadri, Rakotomamonjy, Bach \& Preux}

\RRtitle{Apprentissage de Noyaux \`a Valeurs Op\'erateurs Multiples}

\RRetitle{Multiple Operator-valued Kernel Learning} 

\titlehead{Multiple Operator-valued Kernel Learning}

\RRresume{Dans cet article, nous proposons une m\'ethode d'apprentissage de noyaux multiples \`a valeurs op\'erateurs 
dans le cas d'une r\'egression ridge \`a r\'eponse fonctionnelle. Notre m\'ethode est bas\'ee sur la résolution 
d'un système d'équations linéaires d'opérateurs en utilisant une proc\'edure de type Iterative Coordinate Descent.
Nous validons exp\'erimentalement notre approche 
sur un problème de pr\'ediction de mouvement de doigt dans un contexte d'Interface Cerveau-Machine.}

\RRabstract{Positive definite operator-valued kernels generalize the well-known notion of reproducing kernels, 
and are naturally adapted to multi-output learning situations.
This paper addresses the problem of learning a finite linear 
combination of infinite-dimensional operator-valued kernels 
which are suitable for extending functional data analysis methods to nonlinear contexts.
We study this problem in the case of kernel ridge regression for
functional responses with an $\ell_r$-norm constraint on the
combination coefficients ($r \geq 1$). The resulting optimization
problem is more involved than those of multiple scalar-valued kernel
learning since operator-valued kernels pose more technical and
theoretical issues.
We propose a multiple operator-valued kernel learning algorithm based
on solving a system of linear operator equations by using a block
coordinate-descent procedure. 
We experimentally validate our approach on a functional regression
task  in the context of finger movement prediction in brain-computer interfaces.} 

\RRmotcle{noyaux \`a valeurs op\'{e}rateurs, apprentissage de noyaux multiples, 
analyse des donn\'{e}es fonctionnelles, espace de Hilbert \`a noyau reproduisant}

\RRkeyword{Operator-valued kernels, multiple kernel learning, nonparametric functional data analysis,
 function-valued reproducing kernel Hilbert spaces.
 }
\RRprojets{SequeL} 
 \RRtheme{\THCog} 
 \RCLille 

\begin{document}
\RRNo{7900}
\makeRR   


\section{Introduction}

During the past decades, a large number of algorithms have been
proposed to deal with learning problems in the case of single-valued
functions (\textit{e.g.}, binary-output function for classification or
real output for regression).  Recently, there has been
considerable interest in estimating vector-valued functions~\cite{Micchelli-2005a, Caponnetto-2008, Carmeli-2010}. 
Much of this interest has arisen from the need
to learn tasks where the target is a complex entity, not a scalar
variable. Typical learning situations include multi-task
learning~\cite{Evgeniou-2005}, functional regression~\cite{Kadri-2010}, and
structured output prediction~\cite{Brouard-2011}.

In this paper, we are interested in the problem of functional regression with functional responses in the context of 
brain-computer interface~(BCI) design. 
More precisely, we are interested in finger movement prediction
from electrocorticographic signals~\cite{Schalk2007}. Indeed, from a set of signals
measuring brain surface electrical activity on $d$ channels during a given
period of time, we want to predict, for any instant of that period
whether a finger is moving or not and the amplitude of the finger flexion. 
Formally,  the problem consists in learning a functional dependency
between a set of $d$ signals and a sequence of labels (a step function indicating whether a finger is moving or not) and
between the same set of signals and 
vector of real values (the amplitude function).  
While, it is clear that this problem can be formalized as 
functional regression problem, from our point of view,
such problem can benefit from the multiple operator-valued kernel learning
framework.  Indeed, for these problems, one of the difficulties arises from
the unknown latency between the signal related to the finger movement
and the actual movement~\cite{Pistohl2008}. Hence, instead of fixing in advance some value for this 
latency in the regression model, our framework allows
to learn it from the data by means of several operator-valued kernels.

If we wish to address 
functional regression problem in the principled framework of
reproducing kernel Hilbert spaces (RKHS), we have to consider RKHSs
whose elements are operators that map a function to another function
space, possibly source and target function spaces being
different. Working in such RKHSs, we are able to draw on the important
core of work that has been performed on scalar-valued and vector-valued RKHSs~\cite{Scholkopf-2002,Micchelli-2005a}. 
Such a functional RKHS framework and associated
operator-valued kernels have been introduced recently~\cite{Kadri-2010,kadri-2011}. 
A basic  question with reproducing kernels is how to build these kernels and what is the optimal kernel choice for a given application. 
In order to overcome the need for choosing a kernel before the learning
process, several works have tried to address
the problem of learning the scalar-valued kernel jointly with the 
decision function~\cite{lanckriet_learninker, Sonnenburg-2006}. Since
these seminal works, many efforts have been carried out in order
to theoretically analyze the kernel learning framework~\cite{cortes00:_gener_bound_for_learn_kernel, Bach-2008} or
in order to provide efficient algorithms~\cite{rakoto-2008, aflalo11:_variab_spars_kernel_learn, kloft11:_lp_norm_multip_kernel_learn}. %
While many works have been devoted to
multiple \emph{scalar-valued} kernel learning, this problem
of kernel learning have been barely investigated for \emph{operator-valued}
kernels.
One motivation of this work is to bridge the gap between multiple kernel learning~(MKL)
and operator-valued kernels by proposing a framework and an algorithm
for  learning a finite linear combination of
operator-valued kernels. 
While each step of the scalar-valued MKL 
framework can be 
extended without major difficulties to operator-valued kernels, technical
challenges arise at all stages because we deal with infinite
dimensional spaces. 
It should be pointed out that in a recent work~\cite{Dinuzzo-2011}, 
the problem of learning the output kernel was formulated 
as an optimization problem over the cone of positive semidefinite matrices, and proposed 
a block-coordinate descent method to solve it. However, 
they did not focus on learning the input kernel. 
In contrast, our multiple operator-valued kernel learning formulation 
can be seen as a way of learning simultaneously input and output kernels, 
although we consider a linear combination of kernels which are fixed in advance.\\[-0.3cm]

In this paper, we make the following contributions:
\begin{itemize}
\itemsep=-0.1pt
\item[]~\vspace{-0.73cm}
 \item we introduce a novel approach to infinite-dimensional multiple operator-valued kernel learning (MovKL) suitable 
 for learning the functional dependencies and interactions between continuous data, 
 
  \item  we extend the original formulation of ridge regression in dual variables to the functional data analysis domain, showing how to perform nonlinear functional regression with functional responses by constructing a linear regression operator in an operator-valued kernel feature space~(Section~\ref{sec:PS}),
 
 \item we derive a dual form of the MovKL problem with functional ridge regression, and show that a solution
of the related optimization problem exists~(Section~\ref{sec:PS}),

\item we propose a block-coordinate descent algorithm to solve the MovKL optimization problem which involves solving a challenging linear system with a sum of block operator matrices, and show its convergence in the case of compact operator-valued kernels~(Section~\ref{sec:MovKL}),

\item we provide an empirical evaluation of MovKL performance which demonstrates its effectiveness on a BCI dataset~(Section~\ref{sec:Ex}).

\end{itemize}


\section{Problem Setting}
\label{sec:PS}

Before describing the multiple operator-valued kernel learning algorithm 
that we will study and experiment with in this paper, we first review notions 
and properties of reproducing kernel Hilbert spaces with operator-valued kernels, show their connection to learning from multiple response data~(multiple outputs;~see~\cite{Micchelli-2005a} 
for discrete data and~\cite{Kadri-2010} for continuous data), and describe the optimization problem
for learning kernels with functional response ridge regression.

\subsection{Notations and Preliminaries}

We start by some standard notations and definitions used all along the paper. 
Given a Hilbert space~$\mathcal{H}$, $\langle \cdot,\cdot \rangle_\mathcal{H}$ and 
$\| \cdot \|_\mathcal{H}$ refer to its inner product and norm, respectively.
We denote by $\mathcal{G}_{x}$ and $\mathcal{G}_{y}$ the separable real Hilbert spaces of input and 
output functional data, respectively.
In functional data analysis domain, continuous data are generally assumed to belong to 
the space of square integrable functions $L^2$. In this work, we consider that $\mathcal{G}_{x}$ and $\mathcal{G}_{y}$ 
are the Hilbert space $L^2(\Omega)$ which consists 
of all equivalence classes of square integrable functions on a finite set $\Omega$. 
$\Omega$ being potentially different for $\mathcal{G}_{x}$ and $\mathcal{G}_{y}$.
We denote by $\mathcal{F}(\mathcal{G}_{x},\mathcal{G}_{y})$ the vector space of functions $f: \mathcal{G}_{x} \longrightarrow \mathcal{G}_{y}$, 
and by $\mathcal{L(G}_{y})$ the set of bounded linear operators from $\mathcal{G}_{y}$ to $\mathcal{G}_{y}$.

We consider the problem of estimating a function $f$ such that
$f(x_i)=y_i$ when observed functional data $(x_{i},y_{i})_{i=1,\ldots,n} \in (\mathcal{G}_{x},\mathcal{G}_{y}$).
Since $\mathcal{G}_{x}$ and $\mathcal{G}_{y}$ are spaces of functions, the
problem can be thought of as an operator estimation problem, where the
desired operator maps a Hilbert space of factors to a Hilbert space of
targets. We can define the regularized operator estimate of $f \in
\mathcal{F}$ as:
%
\begin{equation}
  \label{mp}
  f_{\lambda} \triangleq \arg\min\limits_{f \in \mathcal{F}} \frac{1}{n}\sum\limits_{i=1}^{n}\|y_{i}-f(x_{i})\|_{\mathcal{G}_{y}}^{2}
  +\lambda\|f\|_{\mathcal{F}}^{2}.
\end{equation}
In this work, we are looking for a solution to this minimization
problem in a function-valued reproducing kernel Hilbert space $\mathcal{F}$.
More precisely, we mainly focus on the RKHS $\mathcal{F}$ whose elements are 
continuous linear operators on $\mathcal{G}_{x}$ with values in $\mathcal{G}_{y}$. The continuity 
property is obtained by considering a special class of reproducing kernels 
called Mercer kernels~\cite[Proposition\ 2.2]{Carmeli-2010}. Note that in this case, $\mathcal{F}$ is 
separable since $\mathcal{G}_{x}$ and $\mathcal{G}_{y}$ are separable~\cite[Corollary\ 5.2]{Carmeli-2006}.

We now introduce (function)~$\mathcal{G}_{y}$-valued
reproducing kernel Hilbert spaces and show the correspondence
between such spaces and positive definite (operator)~$\mathcal{L(G}_{y})$-valued kernels.
These extend the traditional properties of scalar-valued kernels.
%
%
%
\begin{de} (function-valued RKHS) \\[0.1cm] 
  A Hilbert space $\mathcal{F}$ of functions from $\mathcal{G}_{x}$ to
  $\mathcal{G}_{y}$ is called a reproducing kernel Hilbert space if
  there is a positive definite $\mathcal{L(G}_{y})$-valued kernel
  $K_{\mathcal{F}}(w,z)$ on $\mathcal{G}_{x} \times \mathcal{G}_{x}$ such that:
\vspace{-0.2cm}
  \begin{enumerate}[i.]
    \item \label{enum1:i} the function $z \longmapsto
      K_{\mathcal{F}}(w,z)g$ belongs to $\mathcal{F},\ \forall z \in
      \mathcal{G}_{x},\ w \in \mathcal{G}_{x},\ g \in \mathcal{G}_{y}$,
    \item \label{enum1:ii} $\forall f \in \mathcal{F}, w \in \mathcal{G}_{x},\ g \in \mathcal{G}_{y}, \ \ 
      \langle f,K_{\mathcal{F}}(w,\cdot)g\rangle _{\mathcal{F}} =
      \langle f(w),g\rangle_{\mathcal{G}_{y}}$ \hspace*{0.1cm} (reproducing property).
  \end{enumerate}
\end{de}
%
%
\begin{de} (operator-valued kernel) \\[0.1cm]
  An $\mathcal{L(G}_{y})$-valued kernel $K_{\mathcal{F}}(w,z)$ on
  $\mathcal{G}_{x}$ is a function
  $K_{\mathcal{F}}(\cdot,\cdot):\mathcal{G}_{x} \times \mathcal{G}_{x}
  \longrightarrow \mathcal{L(G}_{y})$; furthermore:
\vspace{-0.1cm}
  \begin{enumerate}[i.]
    \item $K_{\mathcal{F}}$ is Hermitian if
      $K_{\mathcal{F}}(w,z)=K_{\mathcal{F}}(z,w)^{*}$, where $*$ denotes the adjoint operator, 
    \item $K_{\mathcal{F}}$ is positive definite on $\mathcal{G}_{x}$ if it
      is Hermitian and for every natural number $r$ and all
      $\{(w_{i},u_{i})_{i=1,\ldots ,r}\}\in \mathcal{G}_{x} \times
      \mathcal{G}_{y}$,  $\sum_{i,j} \langle
      K_{_{\mathcal{F}}}(w_{i},w_{j})u_{j},u_{i}\rangle_{\mathcal{G}_{y}} \geq 0$.
  \end{enumerate}
\end{de}
\begin{theo} (bijection between function valued RKHS and  operator-valued kernel) \\
  \label{th:positiveDefiniteImpliesKernel}
  An $\mathcal{L(G}_{y})$-valued kernel $K_{\mathcal{F}}(w,z)$ on
  $\mathcal{G}_{x}$ is the reproducing kernel of some Hilbert space
  $\mathcal{F}$, if and only if it is positive definite.
\end{theo}
The proof of Theorem~\ref{th:positiveDefiniteImpliesKernel} can be found in~\cite{Micchelli-2005a}. 
For further reading on operator-valued kernels and their associated RKHSs, see, e.g.,~\cite{Caponnetto-2008, Carmeli-2006, Carmeli-2010}.



\subsection{Functional Response Ridge Regression in Dual Variables}

We can write the ridge regression with functional responses optimization problem~(\ref{mp}) as follows: 
\begin{eqnarray}
\label{minppp}
\begin{array}{ll}

& \min\limits_{f \in \mathcal{F}} \displaystyle\frac{1}{2}\|f\|_{\mathcal{F}}^{2}
+\frac{1}{2n\lambda} \sum\limits_{i=1}^n \|\xi_i\|_{\mathcal{G}_{y}}^{2} \\
& \text{ with } \xi_i = y_i - f(x_i).
\end{array}
\end{eqnarray}
Now, we introduce the Lagrange multipliers $\alpha_i, i=1,\ldots,n$ which are functional 
variables since the output space is the space of functions $\mathcal{G}_{y}$. 
For the optimization problem~(\ref{minppp}), the Lagrangian multipliers exist and the Lagrangian function is well defined. The method of Lagrange multipliers on Banach spaces, which is a generalization of the classical (finite-dimensional) Lagrange multipliers method suitable to solve certain infinite-dimensional constrained optimization problems, is applied here. For more details, see~\cite{Kurcyusz-1976}. 
Let $\alpha=(\alpha_i)_{i=1,\ldots,n} \in \mathcal{G}_{y}^{n}$ the vector of functions containing 
the Lagrange multipliers, the Lagrangian function is defined as 
\begin{equation}
\label{lagrangian}
 L(f,\alpha,\xi) = \displaystyle\frac{1}{2} \|f\|_{\mathcal{F}}^{2} + \frac{1}{2n\lambda} \|\xi\|_{\mathcal{G}_{y}^{n}}^{2} 
+ \langle \alpha, y - f(x) - \xi \rangle_{\mathcal{G}_{y}^{n}},
\end{equation}
where 
$\alpha =  (\alpha_1, \ldots, \alpha_n) \in \mathcal{G}_{y}^{n}$, 
$y =  (y_1, \ldots, y_n) \in \mathcal{G}_{y}^{n}$, 
$\xi =  (\xi_1, \ldots, \xi_n) \in \mathcal{G}_{y}^{n}$,
$f(x) =  (f(x_1), \ldots, f(x_n)) \in \mathcal{G}_{y}^{n}$, 
%
%
and $\forall a, b \in \mathcal{G}_{y}^{n}$, $\langle a , b\rangle_{\mathcal{G}_{y}^{n}} 
= \sum\limits_{i=1}^n \langle a_i , b_i \rangle_{\mathcal{G}_{y}}$.

Differentiating~(\ref{lagrangian}) with respect to $f  \in \mathcal{F}$ and setting to zero, we obtain 
\begin{equation}
 f(.) = \sum\limits_{i=1}^n K(x_i,.)\alpha_i,
\end{equation}
where $K: \mathcal{G}_{x} \times \mathcal{G}_{x} \longrightarrow \mathcal{L(G}_{y})$ 
is the operator-valued kernel of $\mathcal{F}$. 

Substituting this into~(\ref{lagrangian}) and minimizing with respect to $\xi$, we obtain the dual of the functional response ridge regression problem
\begin{equation}
\label{minpdual}
\max\limits_\alpha - \displaystyle\frac{n\lambda }{2}\|\alpha\|_{\mathcal{G}_{y}^{n}}^{2}
- \displaystyle\frac{1}{2}\langle \mathbf{K} \alpha, \alpha \rangle_{\mathcal{G}_{y}^{n}} 
+ \langle \alpha, y \rangle_{\mathcal{G}_{y}^{n}},
\end{equation}
where $\mathbf{K} = [K(x_i,x_j)]_{i,j=1}^n$ is the block operator kernel matrix. 


\subsection{MovKL in Dual Variables}

Let us now consider that the function $f(\cdot)$ is sum
of $M$ functions $\{f_k(\cdot)\}_{k=1}^M$ where each $f_k$ belongs to a
$\mathcal{G}_{y}$-valued RKHS with kernel $K_k(\cdot,\cdot)$.  
Similarly to scalar-valued multiple kernel learning, we adopt the convention that $\frac{x}{0} = 0$ if $x=0$ and $\infty$ otherwise,
and we can cast the problem of learning these functions $f_k$ as 
 \begin{eqnarray}
\label{minp}
\begin{array}{ll}
\displaystyle \min_{d \in \mathcal{D}}& \hspace{-0.3cm} \min\limits_{f_k \in \mathcal{F}_k} \displaystyle\sum_{k=1}^M \displaystyle\frac{\|f_k\|_{\mathcal{F}_k}^{2}
}{2d_k}+\frac{1}{2\lambda} \sum\limits_{i=1}^n \|\xi_i\|_{\mathcal{G}_{y}}^{2} \\[0.2cm]
& \hspace{-0.3cm} \text{ with } \xi_i = y_i - \sum_{k=1}^M f_k(x_i),
%
\end{array}
\end{eqnarray}
where $d=[d_1,\cdots,d_M]$, $\mathcal{D} = \{d: \forall k, d_k\geq 0$ and 
$\sum_k d_k^r \leq 1\} $ and $1\leq r \leq \infty$. Note that this problem
can equivalently be rewritten as an unconstrained optimization problem.  
Before deriving the dual of this problem, it can be shown by means
of the generalized Weierstrass theorem \cite{kurdila05:_convex_funct_analy} that this problem admits a solution (a detailed proof is provided in the
supplementary material).   

Now, following the lines of~\cite{rakoto-2008}, 
a dualization of this problem leads to the following equivalent one
 \begin{equation}
\label{minpmkldual}
\displaystyle \min_{\d \in \mathcal{D}} \max\limits_{\alpha \in \mathcal{G}_{y}^{n}}
 - \displaystyle\frac{n\lambda}{2} \|\alpha\|_{\mathcal{G}_{y}^{n}}^{2}
- \displaystyle\frac{1}{2} \langle \mathbf{K} \alpha, \alpha \rangle_{\mathcal{G}_{y}^{n}} 
+  \langle \alpha, y \rangle_{\mathcal{G}_{y}^{n}}, 
%
\end{equation}
%
%
where $\mathbf{K}= \sum\limits_{k=1}^M d_k \mathbf{K}_k$ and $\mathbf{K}_k$ is the block operator kernel matrix 
associated to the operator-valued kernel $K_k$. The KKT conditions also state that at optimality we have
$f_k(\cdot)= \sum\limits_{i=1}^n d_k K_k(x_i,\cdot)\alpha_i$. 


\section{Solving the MovKL Problem}
\label{sec:MovKL}

After having presented the framework, we now devise an algorithm
for solving this MovKL problem.

\subsection{Block-coordinate descent  algorithm}

Since the optimization problem~(\ref{minp}) has the same structure as
a multiple scalar-valued kernel learning problem, we can build our MovKL algorithm
upon the MKL literature. Hence, we propose to borrow
from~\cite{kloft11:_lp_norm_multip_kernel_learn}, and consider a block-coordinate descent
method. 
The convergence of  a block coordinate descent algorithm which is related closely to 
the Gauss-Seidel method was studied in works of~\cite{Tseng-2001} and others. 
The difference here is that we have operators and block operator matrices rather 
than matrices and block matrices, but this doesn't increase the complexity if 
the inverse of the operators are computable (typically analytically or by spectral decomposition).
Our algorithm iteratively solves the problem with respect
to $\alpha$ with $d$ being fixed and then with respect
to $d$ with $\alpha$ being fixed (see Algorithm~1). After having initialized $\{d_k\}$ to non-zero values, this boils down to the
following steps~:
\vspace{-0.2cm}
\begin{enumerate}
\setlength{\itemsep}{0.2cm}
\item with $\{d_k\}$ fixed, the resulting optimization problem with
respect to $\alpha$ has  a simple closed-form solution:
\begin{equation}
\label{sloe}
  (\mathbf{K} + \lambda I)\alpha = 2 y,
\end{equation}
where $\mathbf{K}= \sum_{k=1}^M d_k \mathbf{K}_k$.
While the form of solution is rather simple, solving this
linear system is still challenging in the operator setting and we propose below an algorithm
for its resolution. 
\item with $\{f_k\}$ fixed, according to problem~(\ref{minp}), we can rewrite
the problem as   
 \begin{eqnarray}
\label{mind}
\begin{array}{ll}
\displaystyle \min_{d \in \mathcal{D}}\sum_{k=1}^M \displaystyle\frac{\|f_k\|_{\mathcal{F}_k}^{2}}{d_k}
\end{array}
\end{eqnarray}
which has a closed-form solution and for which optimality
occurs at~\cite{micchelli_learningkernel}:
\begin{equation}
\label{eq:d}
d_k=\displaystyle\frac{\|f_k\|^{\frac{2}{r+1}}}{(\sum_k\|f_k\|^{\frac{2r}{r+1}})^{1/r}}.
\end{equation}
\end{enumerate}
This algorithm is similar to that of~\cite{Cortes-2009} and~\cite{kloft11:_lp_norm_multip_kernel_learn} 
both being based on alternating optimization. The difference here is that 
we have to solve a linear system involving a block-operator kernel matrix  
which is a combination of basic 
kernel matrices associated to $M$ operator-valued kernels. This
makes the  system very challenging, and we
present an algorithm for solving it in the next paragraph. 

A detailed proof of convergence of the MovKL algorithm, in the case of compact operator-valued kernels, is given in the
supplementary material. It proceeds by showing that the sequence of
functions $\{f_k^{(n)}\}$ generated by the above alternating
optimization produces a non-increasing sequence of objective values of
Equation (\ref{minp}).  Then, using continuity and boundedness
arguments, we can also show that the sequence $\{f_k^{(n)}\}$ is
bounded and thus converges to a minimizer of Equation (\ref{minp}).
The proof is actually an extension of results obtained
by~\cite{Argyriou-2008} and~\cite{Rakoto-2011} for scalar-valued kernels.
However, the extension is not straighforward since infinite-dimensional Hilbert spaces with operator-valued reproducing kernels raise some technical issues that 
we have leveraged in the case of compact operators.

\begin{table}[t]
\small
 \begin{minipage}[t]{1\linewidth}
\begin{tabular}{l}
  \hline \\[-1.0em]
  \textbf{Algorithm 1} \ $\ell_r$-norm MovKL \hspace{7cm} \\
  \hline\\[-0.5em]
  \textbf{Input\ \ \ }  $\mathbf{K}_k$ \ for $k=1,\ldots,M$ \\[0.1cm]
  $d_k^1 \longleftarrow \displaystyle\frac{1}{M}$ \ \ for $k=1,\ldots,M$ \\[0.1cm]
  $\alpha \longleftarrow 0$ \\[0.1cm]
  \textbf{for} $t=1,2,\ldots$ \ \textbf{do} \\[0.1cm]
  $\quad \alpha' \longleftarrow \alpha$ \\[0.1cm]
  $\mathbf{\quad K}  \longleftarrow \sum_k d_k^t \mathbf{K}_k$ \\[0.15cm]
 $\quad \alpha  \longleftarrow$ solution of $(\mathbf{K} + \lambda I) \alpha = 2  y$  \\[0.1cm]
    $\quad$\textbf{if } $\|\alpha-\alpha'\| < \epsilon$ \textbf{ then}\\[0.1cm]
      $\qquad$ break \\[0.1cm]
  $\quad$\textbf{end if}\\[0.1cm]
     $\quad d_k^{t+1} \longleftarrow \displaystyle\frac{\|f_k\|^{\frac{2}{r+1}}}{(\sum_k\|f_k\|^{\frac{2r}{r+1}})^{1/r}}$  \ \ for $k=1,\ldots,M$ \\[0.1cm]
\textbf{end for}\\[0.1cm]
  \hline
\end{tabular}
%
\end{minipage}
\end{table}

\subsection{Solving a linear system with multiple block operator-valued kernels}

One common way to construct operator-valued kernels is to build scalar-valued ones which are carried over to the 
vector-valued (resp. function-valued) setting by a positive definite matrix (resp. operator). In this setting an 
operator-valued kernel has the following form:
\begin{equation*}
 K(w,z) = G(w,z) T,
\end{equation*}
where $G$ is a scalar-valued kernel and $T$ is an operator in
$\mathcal{L(G}_{y})$. In multi-task learning, $T$ is a finite
dimensional matrix that is expected to share information between
tasks~\cite{Evgeniou-2005, Caponnetto-2008}.  More recently and for supervised
functional output learning problems, $T$ is chosen to be a
multiplication or an integral
operator~\cite{Kadri-2010,kadri-2011}. This choice is motivated by the
fact that functional linear models for functional
responses~\cite{Ramsay-2005} are based on these operators and then such kernels provide
an interesting alternative to extend these models to nonlinear
contexts. In addition, some works on functional regression and
structured-output learning consider operator-valued kernels
constructed from the identity operator as in~\cite{Lian-2007} and~\cite{Brouard-2011}.  
In this work we adopt a functional data analysis
point of view and then we are interested in a finite combination of
operator-valued kernels constructed from the identity, multiplication
and integral operators.  A problem encountered when working with
operator-valued kernels in infinite-dimensional spaces is that of
solving the system of linear operator equations~(\ref{sloe}).  In the
following we show how to solve this problem for two cases of
operator-valued kernel combinations.

\textbf{Case 1: multiple scalar-valued kernels and one operator.} \ This is the simpler case where the combination of 
operator-valued kernels has the following form
\begin{equation}
  K(w,z) = \sum\limits_{k=1}^M d_k G_k(w,z)T,
\end{equation}
where $G_k$ is a scalar-valued kernel, $T$ is an operator in $\mathcal{L(G}_{y})$, and $d_k$ are the combination coefficients.
In this setting, the block operator kernel matrix $\mathbf{K}$ can be expressed as a Kronecker product between the 
multiple scalar-valued kernel matrix $\mathbf{G}=\sum_{k=1}^M d_k \mathbf{G}_k$, where 
$\mathbf{G}_k = [G_k(x_i,x_j)]_{i,j=1}^n$, and the operator~$T$. Thus we can compute 
an analytic solution of the system of equations~(\ref{sloe}) 
by inverting $\mathbf{K}+\lambda I$ using the eigendecompositions of 
$\mathbf{G}$ and $T$ as in~\cite{kadri-2011}. 
%
%


\begin{table}[t]
\small
 \begin{minipage}[t]{1\linewidth}
\begin{tabular}{l}
  \hline \\[-1.0em]
  \textbf{Algorithm 2} \ Gauss-Seidel Method \hspace{6.4cm} \\
  \hline\\[-0.5em]
choose an initial vector of functions $\alpha^{(0)}$\\[0.1cm]
\textbf{repeat} \\[0.1cm]
$\quad$ \textbf{for} $i=1,2,\ldots,n$\\[0.1cm]
$\quad\quad$ $\alpha_i^{(t)} \longleftarrow$ sol. of~(\ref{GS}): \\ $\qquad \qquad\qquad$
$[K(x_i,x_i) +  \lambda I] \alpha_i^{(t)} = s_i $
 \\[0.1cm]
$\quad$ \textbf{end for}\\[0.1cm]

\textbf{until} convergence\\[0.1cm]
  \hline
\end{tabular}

\end{minipage}

\end{table} 


\textbf{Case 2: multiple scalar-valued kernels and multiple operators.}
\ This is the general case where multiple operator-valued kernels are
combined as follows
\begin{equation}
\label{msmo}
  K(w,z) = \sum\limits_{k=1}^M d_k G_k(w,z)T_k,
\end{equation}
where $G_k$ is a scalar-valued kernel, $T_k$ is an operator in $\mathcal{L(G}_{y})$, and $d_k$ are the combination coefficients.
Inverting the associated block operator kernel matrix~$\mathbf{K}$ is not feasible in this case, that is why 
we propose a Gauss-Seidel iterative procedure (see Algorithm 2) to solve the system of linear operator equations~(\ref{sloe}). 
Starting from an initial vector of functions $\alpha^{(0)}$, the idea is to iteratively compute, until a convergence condition 
is satisfied, the functions $\alpha_i$ according to the following expression 
%
\begin{equation}
\label{GS}
  [K(x_i,x_i) + \lambda I] \alpha_i^{(t)}  =  2 y_i - \sum\limits_{j=1}^{i-1} K(x_i,x_j)\alpha_j^{(t)}  -  \sum\limits_{j=i+1}^{n} K(x_i,x_j)\alpha_j^{(t-1)},
\end{equation}
where $t$ is the iteration index. This problem is still challenging because the kernel
$K(\cdot,\cdot)$ still involves a positive combination of operator-valued
kernels.  Our algorithm is based on the idea  that instead of inverting the finite combination of operator-valued 
kernels $[K(x_i,x_i) + \lambda I]$, we can consider the following variational formulation of this 
system
\begin{equation*}
 \min\limits_{\alpha_i^{(t)}} \ \frac{1}{2} \ \langle  \sum\limits_{k=1}^{M+1} K_k(x_i,x_i) \alpha_i^{(t)} , \alpha_i^{(t)} \rangle_{\mathcal{G}_{y}}
- \langle s_i, \alpha_i^{(t)} \rangle_{\mathcal{G}_{y}},
\end{equation*}
where $ s_i = 2 y_i - \sum\limits_{j=1}^{i-1} K(x_i,x_j)\alpha_j^{(t)} - \sum\limits_{j=i+1}^{n} K(x_i,x_j)\alpha_j^{(t-1)}$, 
$K_k = d_k G_k T_k$, $\forall k\in\{1,\ldots,M\}$, and  $K_{M+1}=\lambda I$.

Now, by means
of a variable-splitting approach, we are able to decouple the role of the different kernels. Indeed, the above
problem is equivalent to the following one~:
\begin{align*}
  & \min\limits_{\boldsymbol{\alpha}_i^{(t)}} \ \frac{1}{2} \ \langle  \hat {\mathbf{K}}(x_i,x_i) \boldsymbol{\alpha}_i^{(t)} , \boldsymbol{\alpha}_i^{(t)} \rangle_{\mathcal{G}_{y}^M}
- \langle \mathbf{s_i}, \boldsymbol{\alpha}_i^{(t)} \rangle_{\mathcal{G}_{y}^M} \\
 & \text{\ \ \ with \ \ } \alpha_{i,1}^{(t)} = \alpha_{i,k}^{(t)} \ \text{\ \ for \ \ } k=2,\ldots,M+1,
\end{align*}
where $ \hat {\mathbf{K}}(x_i,x_i)$ is the $(M+1) \times (M+1)$ diagonal matrix $[K_k(x_i,x_i)]_{k=1}^{M+1}$. 
$\boldsymbol{\alpha}_i^{(t)}$ is the vector $(\alpha_{i,1}^{(t)},\ldots,\alpha_{i,M+1}^{(t)})$ and 
the $(M+1)$-dimensional vector $\mathbf{s_i} = (s_i,0,\ldots,0)$. We now
have to deal with a quadratic optimization problem with equality constraints. 
Writing down the Lagrangian of this optimization problem and then deriving
its first-order optimality  conditions leads us to the following set of linear equations
\begin{equation}
\label{eq:linsys}
\left\{
\begin{array}{lll}
K_1(x_i,x_i) \alpha_{i,1} - s_i + \sum_{k=1}^{M}\gamma_k & = &  0 \\ 
K_k(x_i,x_i) \alpha_{i,k}  - \gamma_k & = &  0  \\
\alpha_{i,1}-\alpha_{i,k} & = & 0 \\
\end{array}
\right.
\end{equation}
where $k=2,\ldots,M+1$ and $\{\gamma_k\}$ are the Lagrange multipliers related to the 
$M$ equality constraints. Finally, in this set of equations, the 
operator-valued kernels have been decoupled and thus, if their inversion
can be easily computed (which is the case in our
experiments), one can solve the problem~(\ref{eq:linsys}) with respect
to $\{\alpha_{i,k}\}$ and $\gamma_k$  by means of another Gauss-Seidel algorithm after simple reorganization of the linear system.


\section{Experiments}
\label{sec:Ex}


\begin{figure*}[t]
  \centering
~\hfill
  \includegraphics[height=2.7cm, width=3.8cm]{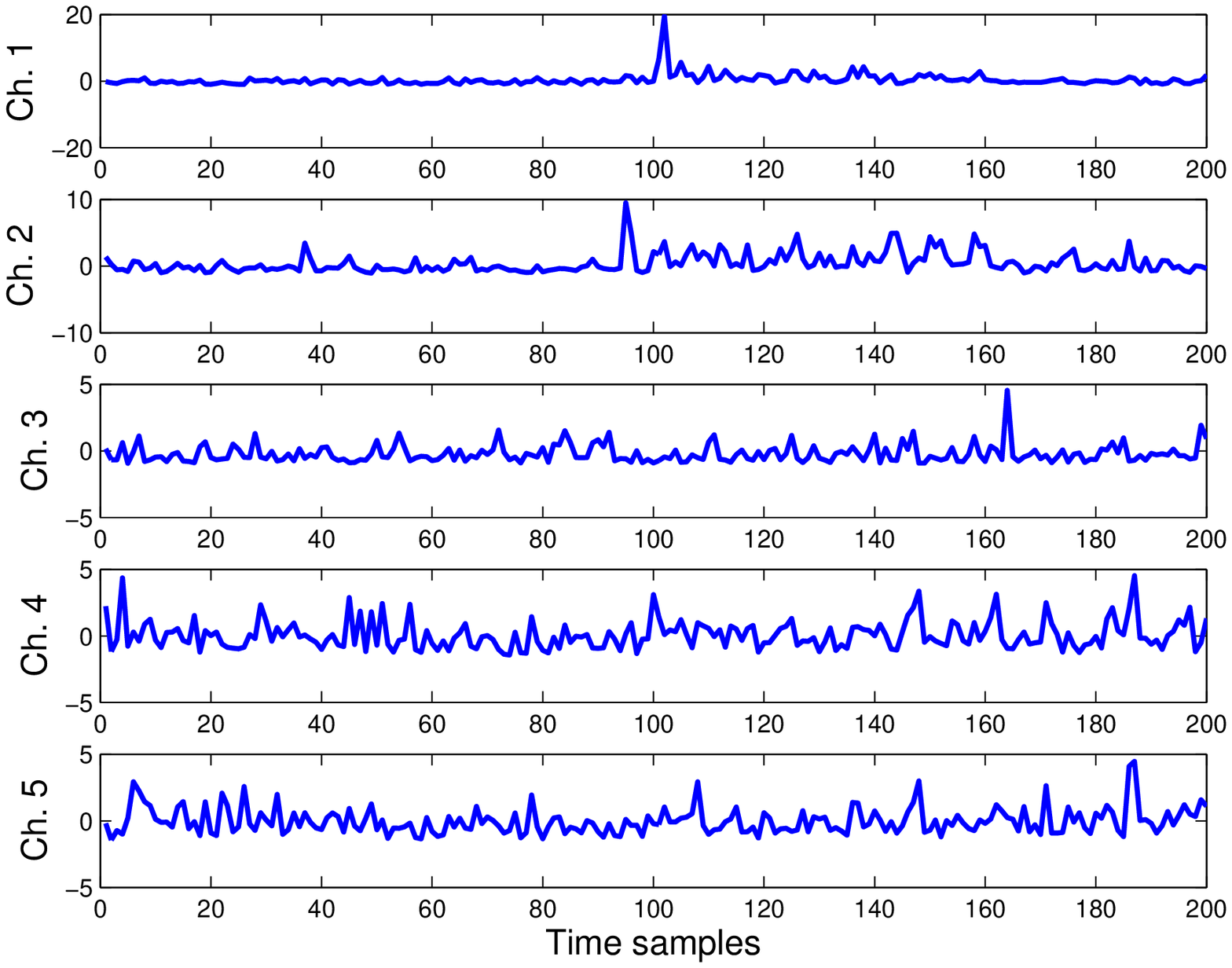} 
\hfill 
\includegraphics[height=2.7cm, width=3.8cm]{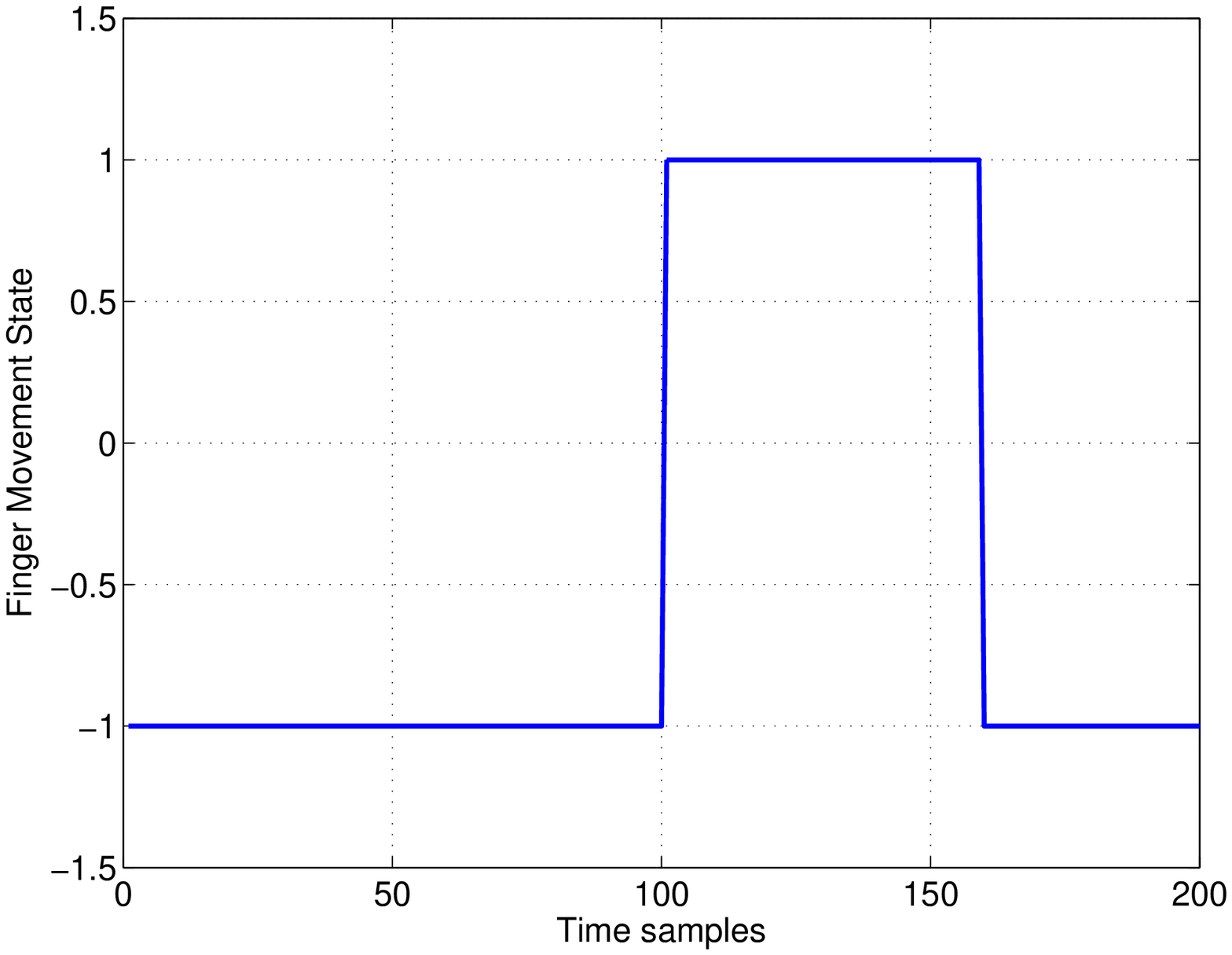}
\hfill~ 
\includegraphics[height=2.7cm, width=3.8cm]{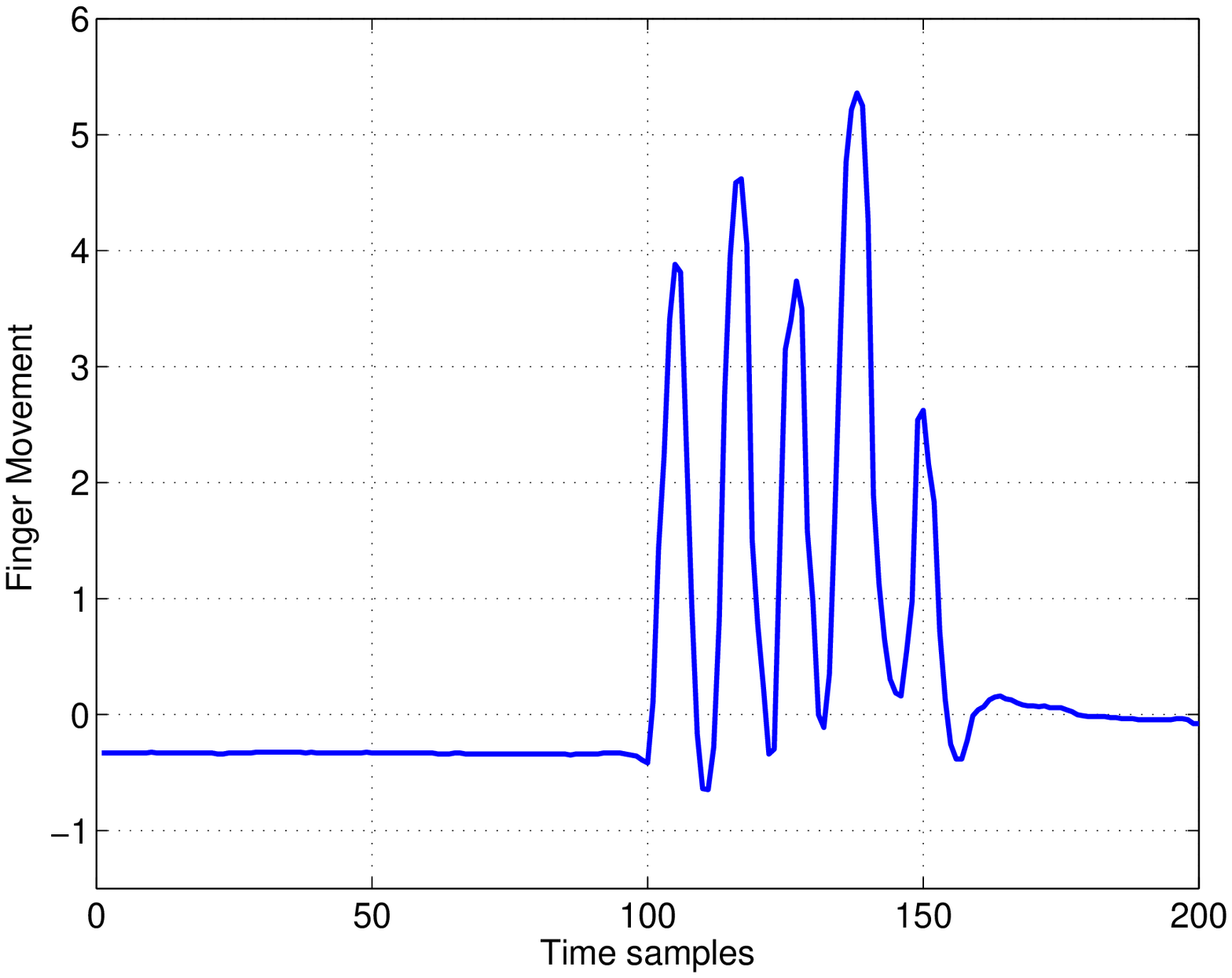}
\hfill~
 \caption{Example of a couple of input-output signals in our BCI task. (left)
Amplitude modulation features extracted from ECoG signals over $5$ pre-defined
channels. 
(middle) Signal of labels denoting whether the finger is moving or
not. (right) Real amplitude movement of the finger.}
  \label{fig:example}
\end{figure*}


In order to highlight the benefit of our multiple operator-valued kernel
learning approach, we have considered a series of experiments on
a real dataset, involving functional output prediction in a brain-computer
interface framework. 
The problem we addressed is a sub-problem related to finger movement
decoding from Electrocorticographic (ECoG) signals. 
We  focus on the problem of estimating if a finger is moving or
not and also on the direct estimation of the finger movement amplitude 
from the ECoG signals.
 The development of the full BCI application is beyond the scope of this
paper and our objective here is to prove that this problem
of predicting finger movement can benefit from multiple kernel learning. 

To this aim, the fourth dataset from the BCI Competition
IV~\cite{bcicompiv} was used. The subjects were~3 epileptic patients
who had platinium electrode grids placed on the surface of their
brains. The number of electrodes varies between 48 to 64 depending on the
subject, and their position on the cortex was unknown.
ECoG\index{ECoG} signals of the subject were
recorded at a 1KHz sampling using BCI2000~\cite{BCI2000}. A band-pass
filter from 0.15 to 200Hz was applied to the ECoG signals. The finger
flexion \index{Finger Flexion} of the subject was recorded at 25Hz and
up-sampled to 1KHz by means of a data glove which measures the
finger movement amplitude. Due to the acquisition process, a delay appears
between the finger movement and the measured ECoG
signal~\cite{bcicompiv}. One of our hopes is that this time-lag
can be properly learnt by means of multiple operator-valued
kernels. Features from the ECoG signals are built by computing some 
band-specific amplitude modulation features, 
which is defined as the sum of the square
of the band-specific filtered ECoG signal samples during a fixed time window.

For our finger movement prediction task, we have kept $5$ channels 
that have been manually selected and split ECoG signals in portions
of $200$ samples. For each of these time segments, we have the label
of whether at each time sample, the finger is moving or not 
as well as the real movement amplitudes. 
The dataset is composed of $487$ couples of input-output signals, the output
signals being either the binary movement labels or the real amplitude movement.
An example of input-output signals are depicted in Figure~\ref{fig:example}. 
In a nutshell, the problem boils down to be a functional regression task with functional responses. 

To evaluate the performance of the multiple operator-valued kernel
learning approach, we use both: \textbf{(1)}~the percentage of labels
correctly recognized~(LCR) defined by $(W_r/T_n)\times 100\%$, where
$W_r$ is the number of well-recognized labels and $T_n$ the total
number of labels to be recognized; \textbf{(2)}~the residual sum of
squares error~(RSSE) as evaluation criterion for curve prediction
\begin{equation}
 \label{rss}
RSSE = \int\sum\limits_{i}\{y_i(t)-\widehat{y}_i(t)\}^2 dt,
\end{equation}
where $\widehat{y}_i(t)$ is the prediction of the function $y_i(t)$ 
corresponding to real finger movement or the finger movement state.
%

\begin{table}[t]
\begin{center}
\caption{Results for the movement state prediction. Residual Sum of Squares Error (RSSE) and the percentage number of Labels Correctly 
Recognized (LCR) of : (1)~baseline KRR with the Gaussian kernel, (2)~functional response KRR with the integral operator-valued kernel, 
(3)~MovKL with $\ell_\infty$, $\ell_1$ and $\ell_2$-norm constraint.}
\label{tableresults1}
\begin{tabular}{lcc}
\\[0cm]
\multicolumn{1}{c}{\bf Algorithm}  &\multicolumn{1}{c}{\bf RSSE}  &\multicolumn{1}{c}{\bf LCR(\%)}
\\ \hline \\[-0.1cm]
KRR - scalar-valued -      & 68.32   & 72.91 \\
KRR - functional response -            & 49.40  & 80.20 \\
MovKL - $\ell_\infty$ norm -               &  45.44 & 81.34\\
MovKL - $\ell_1$ norm -               &  48.12 & 80.66\\
MovKL - $\ell_2$ norm -             & \textbf{39.36}  & \textbf{84.72}\\
\end{tabular}
\end{center}
\end{table}

For the multiple operator-valued kernels having the form~(\ref{msmo}), 
we have used a Gaussian kernel with 5 different bandwidths and a polynomial kernel of degree 
1 to 3 combined with three operators $T$: identity $T y (t) = y(t)$, 
multiplication operator associated with the function $e^{-t^2}$ defined by $T y (t) = e^{-t^2} y(t)$, 
and the integral Hilbert-Schmidt operator with the kernel $e^{-|t-s|}$ proposed 
in~\cite{kadri-2011}, $T y (t) = \int e^{-|t-s|} y(s) ds$. 
The inverses of these operators can be computed analytically. While the inverses of the 
identity and the multiplication operators are easily and directly computable from 
the analytic expressions of the operators, the inverse of the integral operator 
is computed from its spectral decomposition as in~\cite{kadri-2011}. The number of 
eigenfunctions as well as the regularization parameter $\lambda$ are fixed using 
``one-curve-leave-out cross-validation''~\cite{Rice-1991} with the aim of minimizing 
the residual sum of squares error.

Empirical results on the BCI dataset are summarized in 
Tables~\ref{tableresults1} and~\ref{tableresults2} . The dataset was randomly 
partitioned into $65\%$ training and $35\%$ test sets. We compare our approach in the case of 
$\ell_1$ and $\ell_2$-norm constraint on the combination coefficients with: 
(1)~the baseline scalar-valued kernel ridge regression algorithm by considering each 
output independently of the others, 
(2)~functional response ridge regression using an integral operator-valued kernel~\cite{kadri-2011}, 
(3)~kernel ridge regression with an evenly-weighted sum of operator-valued kernels, 
which we denote by $\ell_\infty$-norm MovKL. 

As in the scalar case, using multiple operator-valued kernels leads to better results. 
By directly combining kernels constructed from identity, multiplication and integral operators we could 
reduce the residual sum of squares error and enhance the label classification accuracy. Best results 
are obtained using the MovKL algorithm with $\ell_2$-norm constraint on the combination coefficients. 
RSSE and LCR of the baseline kernel ridge regression are significantly outperformed by the operator-valued kernel 
based functional response regression. These results confirm that by taking into account the relationship between 
outputs we can improve performance. This is due to the fact that an operator-valued kernel induces a similarity 
measure between two pairs of input/output.


\begin{table}[t]
\caption{
Residual Sum of Squares Error (RSSE) results for finger movement prediction.}
\label{tableresults2}
\begin{center}
\begin{tabular}{lc}
\multicolumn{1}{c}{\bf Algorithm}  &\multicolumn{1}{c}{\bf RSSE}  \\ \hline \\[-0.1cm]
KRR - scalar-valued -      &  88.21  \\
KRR - functional response -            & 79.86 \\
MovKL - $\ell_\infty$ norm -               & 76.52 \\
MovKL - $\ell_1$ norm -               &  78.24 \\
MovKL - $\ell_2$ norm -             & \textbf{75.15}\\
\end{tabular}
\end{center}
  
\end{table}


\section{Conclusion}
\label{sec:Conclusion}

In this paper we have presented a new method for learning simultaneously an operator and a finite linear combination of 
operator-valued kernels. We have extended the MKL framework to deal with functional response kernel ridge regression 
and we have proposed a block coordinate descent algorithm to solve the resulting optimization problem. 
The method is applied on a BCI dataset to predict finger movement in a functional regression setting. 
Experimental results show that our algorithm achieves good performance outperforming existing methods. 
It would be interesting for future work to thoroughly compare the proposed MKL 
method for operator estimation with 
previous related methods for multi-class and multi-label MKL
%
in the contexts of structured-output learning
and collaborative filtering.


\section*{Appendix}

\appendix{
\section{Existence of  Minimizers}

We discuss in this section the existence of minimizers of problems (\ref{mp}) and
(\ref{minp}). Because in both problems, we deal with infinite dimensional spaces in the optimization problem, we have to consider appropriate tools for doing so. 

Existence of $f_\lambda$ in the problem given in Equation (\ref{mp}) is guaranteed, for $\lambda>0$ by the generalized Weierstrass Theorem and one of its corollary that
we both remind below \cite{kurdila05:_convex_funct_analy}.
\begin{theo}
 Let $X$ be a reflexive Banach space and $C \subseteq X$ a weakly closed
 and bounded set. Suppose $h: C \mapsto \mathbb{R}$ is a proper lower
semi-continuous function. Then $h$ is bounded from below and has a minimizer
on $C$. 
\end{theo}

\begin{corrol}
Let $H$ be a Hilbert space and $h : H \mapsto \mathbb{R}$ is a strongly lower semi-continuous, convex and coercive function. Then $h$ is bounded from below and attains a minimizer.
\end{corrol}

We can straighforwadly apply this corollary to Problem (\ref{mp}) 
by defining 
$$h(f) = \sum\limits_{i=1}^{n}\|y_{i}-f(x_{i})\|_{\mathcal{G}_{y}}^{2}
+\lambda\|f\|_{\mathcal{F}}^{2}$$
with $f \in \mathcal{F}$ (which is an Hilbert space). It is easy to note
that $h$ is continuous and convex. Besides, $h$ is coercive for
$\lambda >0$ since $\|f\|^2_{\mathcal{F}}$ is coercive and the sum involves
only positive terms. Hence $f_\lambda$ exists. 

Regarding the MKL problem given in (\ref{minp}), we show existence of
a solution in $\d$ and $\{f_k\}$ by 
defining the 
function, for fixed $\{d_k\}$
$$h_1(f_1,\cdots,f_k;\{d_k\}_k) = \sum\limits_{i=1}^{n}\|y_{i}-\sum_k f_k(x_{i})\|_{\mathcal{G}_{y}}^{2}
+\lambda \sum_k \frac{\|f_k\|_{\mathcal{F}}^{2}}{d_k}$$
and by rewriting problem (\ref{minp}) as
$$
\min_{\d \in \mathcal{D}} J(\d) \quad \text{with} \quad J(\d)= \min_{\{f_k\}} h_1(f_1,\cdot,,f_k;\{d_k\}_k)
$$

Using similar arguments as above, it can be shown that $h_1$ is proper, strictly convex and 
coercive for fixed non-negative $\{d_k\}$ and $\lambda>0$ (remind the convention
that $\frac{x}{0} = 0$ if $x=0$ and $\infty$ otherwise). Hence, minimizers of $h_1$  w.r.t.
$\{f_1,\cdots,f_k\}$ exists and are unique. Since the function 
$J(\d)$ which is equal to $h_1(f_1^\star,\cdots,f_k^\star;\{d_k\}_k)$ is continuous
over the compact subset of $\mathbb{R}^M$ defined by
the constraints on $\d$, it also attains its minimum. This conclude
the proof that a solution of problem~(\ref{minp}) exists.


\section{Dual Formulation of Functional Ridge Regression}

Essential computational details regarding the dual formulation of functional ridge regression presented in Section~\ref{sec:PS} are discussed here. 

The functional response ridge regression optimization problem has the following form:
\begin{eqnarray}
\label{minppp_a}
\begin{array}{ll}

& \min\limits_{f \in \mathcal{F}} \displaystyle\frac{1}{2}\|f\|_{\mathcal{F}}^{2}
+\frac{1}{2n\lambda} \sum\limits_{i=1}^n \|\xi_i\|_{\mathcal{G}_{y}}^{2} \\
& \text{ with } \xi_i = y_i - f(x_i).
\end{array}
\end{eqnarray}
where $(x_{i},y_{i})_{i=1,\ldots,n} \in (\mathcal{G}_{x},\mathcal{G}_{y})$.  $\mathcal{G}_{x}$ and $\mathcal{G}_{y}$ 
are the Hilbert space $L^2(\Omega)$ which consists of all equivalence classes of square integrable functions on a finite set $\Omega$, 
and $\mathcal{F}$ is a RKHS whose elements are continuous linear operators on $\mathcal{G}_{x}$ with values in $\mathcal{G}_{y}$. 
$K$ is the $\mathcal{L(G}_{y})$-valued reproducing kernel of $\mathcal{F}$.

Since $\mathcal{G}_{x}$ and $\mathcal{G}_{y}$ are functional spaces, to derive a ``dual version'' of problem~(\ref{minppp_a}) we use 
the method of Lagrange multipliers on Banach spaces which is suitable to solve certain infinite-dimensional constrained optimization problems. 
The method is a generalization of the classical method of Lagrange multipliers.
The existence of Lagrangian multipliers for the problem~(\ref{minppp_a}) which involves an equality constraint is guaranteed by Theorem~4.1~\footnote{This theorem considers only equality constraint, but it is a particular case of Theorem~3.1 in~\cite{Kurcyusz-1976} which deals with more general context.} in~\cite{Kurcyusz-1976}. As consequence, the Lagrangian function associated to~(\ref{minppp_a}) is well defined and Fr\'echet-differentiable. 
Let $\alpha=(\alpha_i)_{i=1,\ldots,n} \in \mathcal{G}_{y}^{n}$ the vector of functions containing 
the Lagrange multipliers, the Lagrangian function is given by
\begin{equation}
\label{lagrangian_a}
 L(f,\alpha,\xi) = \displaystyle\frac{1}{2} \|f\|_{\mathcal{F}}^{2} + \frac{1}{2n\lambda} \|\xi\|_{\mathcal{G}_{y}^{n}}^{2} 
+ \langle \alpha, y - f(x) - \xi \rangle_{\mathcal{G}_{y}^{n}},
\end{equation}
where 
$\alpha =  (\alpha_1, \ldots, \alpha_n) \in \mathcal{G}_{y}^{n}$, 
$y =  (y_1, \ldots, y_n) \in \mathcal{G}_{y}^{n}$, 
$\xi =  (\xi_1, \ldots, \xi_n) \in \mathcal{G}_{y}^{n}$,
$f(x) =  (f(x_1), \ldots, f(x_n)) \in \mathcal{G}_{y}^{n}$, 
%
%
and $\forall a, b \in \mathcal{G}_{y}^{n}$, $\langle a , b\rangle_{\mathcal{G}_{y}^{n}} 
= \sum\limits_{i=1}^n \langle a_i , b_i \rangle_{\mathcal{G}_{y}}$.

Now, we compute $L'(f)$ the derivative of $L(f,\alpha,\xi)$ with respect to $f$ using the G\^ateaux derivative~(generalization of the directional derivative) which can be defined for the direction $h \in \mathcal{F}$ by:
\begin{equation}
  \nonumber D_{h}L(f) = \lim\limits_{\tau \longrightarrow
    0} \frac{L(f+\tau h) - L(f)}{\tau}
\end{equation}
Using the fact that $D_{h}L(f) = \langle L'(f),h\rangle_\mathcal{F} $, we obtain
\begin{enumerate}[i.]
  \item \label{enum3:i} $G(f)=\|f\|_{\mathcal{F}}^{2}$

    $\lim\limits_{\tau \longrightarrow 0} \displaystyle \frac{\|f+\tau
      h\|_{\mathcal{F}}^{2}-\|f\|_{\mathcal{F}}^{2}}{\tau} = 2\langle f,h\rangle \ $
    $\Longrightarrow \   G^{'}(f)=2f$ \\~
  \item \label{enum3:ii} $H(f)=\langle \alpha, y - f(x) - \xi \rangle_{\mathcal{G}_{y}^{n}}$

 $ \displaystyle \lim\limits_{\tau \longrightarrow
        0} \displaystyle \frac{\langle\alpha, y - f(x) - \tau h(x) - \xi \rangle_{\mathcal{G}_{y}^{n}} - \langle\alpha, y - f(x) - \xi \rangle_{\mathcal{G}_{y}^{n}}}{\tau} 
       = -\langle \alpha, h(x)\rangle_{\mathcal{G}_{y}^{n}}$ 
\vspace{2mm}
      $ = - \sum_i  \langle \alpha_i, h(x_i)\rangle_{\mathcal{G}_{y}} = - \langle \sum_i K_{\mathcal{F}}(x_{i},\cdot)\alpha_i,h\rangle _{\mathcal{F}} \ \ \footnotesize
      \text{(using the reproducing property)}$
\vspace{5mm}
\\
    $\Longrightarrow H^{'}(f)= - \sum_iK_{\mathcal{F}}(x_{i},\cdot)\alpha_{i}$
\end{enumerate}
\vspace{0mm}
(\ref{enum3:i}), (\ref{enum3:ii}), and $L^{'}(f)=0$, we obtain the (representer theorem) solution:
\begin{equation}
 f(\cdot)=\displaystyle
\sum\limits_{i=1}^{n}K_{\mathcal{F}}(x_{i},\cdot)\alpha_{i}
\end{equation}

Substituting this into~(\ref{lagrangian_a}), the problem~(\ref{minppp_a}) becomes
\begin{equation}
\label{minmaxp_a}
 \min\limits_{\xi} \max\limits_{\alpha} -\frac{1}{2} 
\langle \mathbf{K} \alpha, \alpha \rangle_{\mathcal{G}_{y}^{n}}
+ \frac{1}{2n\lambda} \|\xi\|_{\mathcal{G}_{y}^{n}}^{2}
+ \langle \alpha, y-\xi\rangle_{\mathcal{G}_{y}^{n}}
\end{equation}
where $\mathbf{K} = [K(x_i,x_j)]_{i,j=1}^n$ is the block operator kernel matrix. 

Differentiating~(\ref{minmaxp_a}) with respect to $\xi$ using the same procedure as described above, 
we obtain $\xi = n\lambda\alpha$ and then the dual of the functional 
response ridge regression problem is given by
\begin{equation}
\label{minpdual_a}
\max\limits_\alpha - \frac{n\lambda}{2} \|\alpha\|_{\mathcal{G}_{y}^{n}}^{2}
- \frac{1}{2} \langle \mathbf{K} \alpha, \alpha \rangle_{\mathcal{G}_{y}^{n}} 
+ \langle \alpha, y \rangle_{\mathcal{G}_{y}^{n}}
\end{equation}


\section{Convergence of Algorithm 1}

In this section, we present a proof of convergence of Algorithm~1. 
The proof is an extension of results obtained by~\cite{Argyriou-2008} and~\cite{Rakoto-2011} to 
infinite dimensional Hilbert spaces with operator-valued reproducing kernels. 
Let $R(f,d)$ be the objective function of the MovKL problem defined by~(\ref{minp}):
\begin{equation*}
R(f,d) = L + \sum_{k=1}^M \frac{\|f_k\|_{\mathcal{F}_k}^{2}}{d_k}
\end{equation*}
where $L = \frac{1}{\lambda}  \sum\limits_i \|y_i - \sum_{k=1}^M f_k(x_i)\|_{\mathcal{G}_{y}}^{2}$. 
Substituting Equation~(\ref{eq:d}) in $R$ we obtain the objective function:
\begin{equation*}
 S(f) := R(f,d(f)) = L +  \left(\sum_{k=1}^M \|f_k\|_{\mathcal{F}_k}^{\frac{2r}{r+1}}\right)^{\frac{r+1}{r}}
\end{equation*}
The function $S$ is \textbf{strictly convex} since $L$ is convex and the function defined by 
$f \longmapsto  \left(\sum_{k=1}^M \|f_k\|^{\frac{2r}{r+1}}\right)^{\frac{r+1}{r}}$ is strictly 
convex (this follows directly from strict convexity of the function $x \longmapsto x^p$ when $x \geq 0$ and $p>1$). 
Thus, $S(f)$ admits a \textbf{unique minimizer}.

Now let us define the function $g$ by:
\begin{equation*}
 g(f) = \underset{u}\min\{R(u,d(f))\}.
\end{equation*}
The function $g$ is \textbf{continuous}. This comes from the fact that the function:
\begin{equation*}
 G(d) =  \underset{u}\min\{R(u,d)\}
\end{equation*}
is continuous. Indeed, $G$ is the minimal of value of a functional response kernel ridge regression problem in 
a function-valued RKHS associated to an operator-valued kernel $K$. So, $G(d) = R(d,u^*)$ with 
$u^* = \big(\mathbf{K}(d) + \lambda I \big)^{-1} y$~(see Equation~(\ref{sloe})). 
$u^*$ is continuous, and hence $G(d)$ is also continuous.

By definition we have $S(f) = R(f,d(f))$, and since $d(f)$ minimizes $R(f,\cdot)$, we obtain that:
\begin{equation*}
 S(f^{(n+1)}) \leq g(f^{(n)}) \leq S(f^{(n)})
\end{equation*}
where $n$ is the number of iteration. So, the sequence $\{S(f^{(n)}), \ n\in \mathbb{N}\}$ is nonincreasing 
and then it is bounded since $L$ is bounded from below. Thus, as $n \longrightarrow \infty$, 
$S(f^{(n)})$ converges to a number which we denote by $S^*$. $\{S(f^{(n)})\}$ is convergent and $S$ is a coercive function, 
then the sequence $\{\|f^{(n)}\|, \ n\in \mathbb{N}\}$ is bounded. Consequently, the sequence 
$\{f^{(n)}, \ n\in \mathbb{N}\}$ is \textbf{bounded}.

Next we show the following subsequence convergence property which underlies the convergence of 
Algorithm~1.
\begin{prop}
\label{prop:subsequence}
 If $\mathcal{F}$ is a RKHS associated to a compact-operator-valued kernel, the sequence $\{f^{(n)} \in\mathcal{F} , \ n\in \mathbb{N}\}$, since it is bounded, has a convergent subsequence.
\end{prop}
\textit{Proof.} The analogue of the Bolzano-Weierstrass 
theorem\footnote{The Bolzano-Weierstrass theorem states that each bounded sequence in $\mathbb{R}^n$ has a convergent subsequence. 
For infinite-dimensional spaces, strong convergence of the subsequence is not reached and only weak convergence is obtained. 
Proposition~\ref{prop:subsequence} shows that strong convergence can be reached for the sequence $\{f^{(n)}, \ n\in \mathbb{N}\}$ solution of our MovKL optimization problem in Hilbert spaces with reproducing compact operator-valued kernels.} in Hilbert spaces states that there exists a \textbf{weakly convergent} subsequence  
$\{f^{(n_l)}, \ l\in \mathbb{N}\}$ of the bounded sequence $\{f^{(n)}\}$ which (weakly) converges to $f \in \mathcal{F}$. 
By definition of weakly convergence, we have $\forall g \in \mathcal{F}$ 
($\mathcal{F}$ is a RKHS with the operator-valued kernel $K$):
 \begin{equation}
 \label{eq:WeakConv}
\underset{n_l \rightarrow \infty}\lim \langle f^{(n_l)}(\cdot), g \rangle_{\mathcal{F}} = \langle f(\cdot), g \rangle_{\mathcal{F}}  
 \end{equation}
 Let $g = K(x,\cdot)\beta$. Using the reproducing property, we obtain
\begin{align*}
& \underset{n_l \rightarrow \infty}\lim \langle f^{(n_l)}(\cdot), g \rangle_{\mathcal{F}} = \underset{n_l \rightarrow \infty}\lim \langle f^{(n_l)}(x), \beta \rangle_{\mathcal{G}_{y}} \quad \text{and} \quad \langle f(\cdot), g \rangle_{\mathcal{F}} = \langle f(x), \beta \rangle_{\mathcal{G}_{y}} \\
& \Rightarrow  \underset{n_l \rightarrow \infty}\lim \langle f^{(n_l)}(x), \beta \rangle_{\mathcal{G}_{y}} =  \langle f(x), \beta \rangle_{\mathcal{G}_{y}}  \quad \text{using~(\ref{eq:WeakConv})}
\end{align*}
Thus the subsequence $\{f^{(n_l)}(\cdot)\}$ is (weakly) \textbf{pointwise convergent}.

Now we show that: 
\begin{equation*}
 \underset{n_l \rightarrow \infty}\lim \|f^{(n_l)}(\cdot)\|_{\mathcal{F}}  = \|f(\cdot)\|_{\mathcal{F}}  \tag{\text{$*$}}
\end{equation*}
Since $f^{(n_l)} \in \mathcal{F}$ is solution of the minimization of the optimization problem~(\ref{minp}) with the kernel combination coefficients $d_k$ fixed, it can be written as $\sum\limits_i K(x_i,\cdot)\alpha_i^{(n_l)}$ (representer theorem). 
We now that $f^{(n_l)}(x)$ converges weakly to $f(x)$, so $\alpha^{(n_l)} = \left(\alpha_i^{(n_l)}\right)_{i \geq 1}$ converges weakly to $\alpha \in \mathcal{G}_{y}^{n}$. Indeed, $\forall \beta \in \mathcal{G}_{y}$ we have:
\begin{align*}
& \underset{n_l \rightarrow \infty}\lim \langle f^{(n_l)}(x), \beta \rangle_{\mathcal{G}_{y}} = \langle f(x), \beta \rangle_{\mathcal{G}_{y}} \\
& \Rightarrow \underset{n_l \rightarrow \infty}\lim \langle \sum_i K(x_i,x)\alpha_i^{(n_l)}, \beta \rangle_{\mathcal{G}_{y}} = \langle \sum_i K(x_i,x)\alpha_i, \beta \rangle_{\mathcal{G}_{y}} \\
& \Rightarrow \underset{n_l \rightarrow \infty}\lim \langle \mathbf{K_x}\alpha^{(n_l)}, \beta \rangle_{\mathcal{G}_{y}} = \langle \mathbf{K_x}\alpha, \beta \rangle_{\mathcal{G}_{y}} \quad \text{where } \mathbf{K_x} \text{ is the row vector } (K(x_i,x))_{i \geq 1} \\
& \Rightarrow \underset{n_l \rightarrow \infty}\lim \langle \alpha^{(n_l)} , \mathbf{K_x^*} \beta \rangle_{\mathcal{G}_{y}^n} = \langle \alpha, \mathbf{K_x^*} \beta \rangle_{\mathcal{G}_{y}^n} \\
& \Rightarrow \underset{n_l \rightarrow \infty}\lim \langle \alpha^{(n_l)} , z \rangle_{\mathcal{G}_{y}^n} = \langle \alpha, z \rangle_{\mathcal{G}_{y}^n} \quad \forall z\in \mathcal{G}_{y}^n \\
& \Rightarrow \alpha^{(n_l)} \text{ converges weakly to } \alpha
\end{align*}

Moreover
\begin{align*}
 \underset{n_l \rightarrow \infty}\lim \|f^{(n_l)}(\cdot)\|_{\mathcal{F}}^2 & = \underset{n_l \rightarrow \infty}\lim \langle \sum_i K(x_i,\cdot)\alpha_i^{(n_l)}, \sum_j K(x_j,\cdot)\alpha_j^{(n_l)} \rangle_{\mathcal{F}}   \\    
 & =  \underset{n_l \rightarrow \infty}\lim \sum_{i,j} \langle K(x_i,x_j)\alpha_i^{(n_l)} , \alpha_j^{(n_l)} \rangle_{\mathcal{G}_{y}} \ \footnotesize \text{(using the reproducing property)} \\
 & =  \underset{n_l \rightarrow \infty}\lim \langle \mathbf{K}\alpha^{(n_l)} , \alpha^{(n_l)} \rangle_{\mathcal{G}_{y}^n} \\
 & =   \langle \mathbf{K}\alpha , \alpha \rangle_{\mathcal{G}_{y}^n}  \quad  \text{(because of the compactness\footnotemark[3] of $\mathbf{K}$)} \\
 & = \sum_{i,j} \langle  K(x_i,x_j)\alpha_i,\alpha_j \rangle_{\mathcal{G}_{y}} \\
& = \sum_{i,j} \langle K(x_i,\cdot)\alpha_i,  K(x_j,\cdot)\alpha_j \rangle_{\mathcal{F}} = \| f \|_{\mathcal{F}}^2
\end{align*}

Using ($*$) and weak convergence, we obtain the \textbf{strong convergence} of the subsequence $\{f^{(n_l)}\}$.
\begin{align*}
 \underset{n_l \rightarrow \infty}\lim \|f^{(n_l)} -f \|_{\mathcal{F}}^2 &= \underset{n_l \rightarrow \infty}\lim \langle f^{(n_l)} -f , f^{(n_l)} \rangle_{\mathcal{F}} - \langle f^{(n_l)} -f , f \rangle_{\mathcal{F}} \\
& = \underset{n_l \rightarrow \infty}\lim \|f^{(n_l)} \|_{\mathcal{F}}^2  - \underset{n_l \rightarrow \infty}\lim 2 \langle f, f^{(n_l)} \rangle_{\mathcal{F}} +  \|f\|_{\mathcal{F}}^2 \\
& = \underset{n_l \rightarrow \infty}\lim \|f^{(n_l)} \|_{\mathcal{F}}^2 - \|f\|_{\mathcal{F}}^2 \ \footnotesize \text{(using weak convergence)}  \\
& = 0 \quad \text{(using($*$))} \\ 
& \Rightarrow f^{(n_l)} \text{ converges strongly to } f \qquad \square
\end{align*}
\footnotetext[3]{Compact operator maps weakly convergent sequences into strongly convergent sequences.}

By Proposition~\ref{prop:subsequence}, there exists a convergent subsequence $\{f^{(n_l)}, \ l\in \mathbb{N}\}$ 
of the bounded sequence $\{f^{(n)}, \ n\in \mathbb{N}\}$, whose limit we denote by $f^*$. Since 
$ S(f^{(n+1)}) \leq g(f^{(n)}) \leq S(f^{(n)})$ , $g(f^{(n)})$ converges to $S^*$. Thus, by the continuity of 
$g$ and $S$, $g(f^*) = S(f^*)$. This implies that $f^*$ is a minimizer of $R(\cdot,d(f^*))$, because 
$R(f^*,d(f^*)) = S(f^*)$. Moreover, $d(f^*)$ is the minimizer of $R(\cdot,f^*)$ subject to the constraints on $d$. 
Thus, since the objective function $R$ is smooth, the pair $(f^*,d(f^*))$ is the minimizer of $R$.

At this stage, we have shown that any convergent subsequence of $\{f^{(n)}, \ n\in \mathbb{N}\}$ converges to 
the minimizer of~$R$. Since the sequence $\{f^{(n)}, \ n\in \mathbb{N}\}$ is bounded, it follows that the whole 
sequence \textbf{converges} to minimizer of~$R$.

}



\bibliographystyle{plain}

\bibliography{MKL_OVK,movklAR}
\end{document}